\pdfoutput=1
\documentclass[11pt]{article}

\usepackage[final]{acl}

\usepackage{times}
\usepackage{latexsym}

\usepackage[T1]{fontenc}
\usepackage[utf8]{inputenc}

\usepackage{enumitem}
\usepackage{microtype}

\usepackage{inconsolata}

\usepackage{xspace}
\usepackage{graphicx}
\usepackage{amsmath}
\usepackage{amssymb}
\usepackage{hyperref}
\usepackage{enumitem}
\usepackage{booktabs}
\usepackage{multirow}
\usepackage{fancyvrb}
\usepackage{xcolor}
\usepackage[colorinlistoftodos]{todonotes}

\newcommand{\eg}{e.g.,\xspace}
\newcommand{\ie}{i.e.,\xspace}

\def\tabref#1{Table~\ref{tab:#1}}

\def\tablabel#1{\label{tab:#1}\label{p:#1}}
\def\eqref#1{Eq.~(\ref{eqn:#1})}
\def\eqlabel#1{\label{eqn:#1}}
\def\Secref#1{Section~\ref{sec:#1}}
\def\Appref#1{Appendix~\ref{sec:#1}}
\def\seclabel#1{\label{sec:#1}\label{p:#1}}

\def\genericname{\mbox{{MaskLID}}\xspace}
\def\genericnamenospace{\mbox{{MaskLID}}}
\def\fasttext{\mbox{FastText}\xspace}

\title{\genericname: Code-Switching Language Identification through Iterative Masking}

\author{Amir Hossein Kargaran$^{\clubsuit}$, François Yvon$^{\spadesuit}$ and Hinrich Schütze$^{\clubsuit}$ \\
        $^\clubsuit$LMU Munich \& Munich Center for Machine Learning, Munich, Germany \\
        $^\spadesuit$Sorbonne Université \& CNRS, ISIR, Paris, France \protect \\
        \texttt{amir@cis.lmu.de}
}

\begin{document}
\maketitle

\begin{abstract}
We present \genericname, a simple, yet effective, code-switching (CS) language identification (LID) method. \genericname does not require any training and is designed to complement current high-performance sentence-level LIDs. Sentence-level LIDs are classifiers trained on monolingual texts to provide single labels, typically using a softmax layer to turn scores into probabilities. However, in cases where a sentence is composed in both L1 and L2 languages, the LID classifier often only returns the dominant label L1. To address this limitation, \genericname employs a strategy to mask text features associated with L1, allowing the LID to classify the text as L2 in the next round. This method uses the LID itself to identify the features that require masking and does not rely on any external resource. In this work, we explore the use of \genericname for two open-source LIDs (GlotLID and OpenLID), that are both based on the \fasttext architecture. Code and demo are available at \href{https://github.com/cisnlp/MaskLID}{\path{github.com/cisnlp/MaskLID}}.

\end{abstract}

\section{Introduction \label{sec:introduction}}

Code-switching (CS), the juxtaposition of two or more languages within a single discourse \citep{gumperz1982discourse}, is prevalent in both written and spoken communication \citep{sitaram2019survey, dogruoz-etal-2021-survey}. While CS has traditionally been explored as a speech phenomenon~\citep{milroy1995one,auer2013code}, the increasing prevalence of CS in digital communication, such as SMS and social media platforms~\citep{das-gamback-2013-code,bali-etal-2014-borrowing}, requires the development of techniques to also analyze CS in written texts. There is however a lack of CS data for researchers, making it difficult to study CS and to effectively train CS-aware models. This shortage affects many NLP applications dealing with CS scenarios~\citep{calcs-2021-approaches,calcs-2023-approaches}. A first step towards the collection of high-quality corpora of CS texts is thus to identify samples of CS in running texts.

Previous works on CS language identification (LID) have mainly focused on building \emph{word-level} LIDs for code-switching between specific pairs of languages, and are often limited to recognize only two languages~\citep{solorio-etal-2014-overview, nguyen-dogruoz-2013-word, elfardy2013code, barman-etal-2014-code}. However, such approaches are not realistic on a larger scale, especially considering that texts on the web typically lack prior information about the languages that are actually being used.

More recently, \citet{burchell-etal-2024-code} have investigated the use of high-quality LID at the \emph{sentence-level} to detect instances of CS. They propose to reformulate CS LID as a \emph{sentence-level} task and to associate each segment with \emph{a set of language labels}. Their investigation reveals the difficulty of achieving effective CS LID with existing LID models. Furthermore, their findings indicate that such LIDs predominantly predict only one of the languages occurring in CS sentences.

In this work, we continue this line of reserach and introduce \genericname, a method that also uses high-quality sentence-level LID to identify CS segments. By masking the presence of the text features associated with the dominant language, \genericname improves the ability to recognize additional language(s) as well. We explain in detail how \genericname works in cooperation with two existing LIDs that are based on the \fasttext \cite{bojanowski-etal-2017-enriching} architecture in \Secref{method}. As we discuss, our method can identify arbitrary pairs of languages, and is also able to detect mixtures of more than two languages in the same segment. Being based on \fasttext{}, it is also extremely fast. This two properties make \genericname{} well suited to mine large web corpora for examples of real-world CS segments, that can then serve as valuable training data for applications designed to handle CS inputs. We evaluate \genericname on two test datasets containing both CS and monolingual data, showing the benefits of using \genericname (see \Secref{results}). 

\section{One Sentence, Multiple Languages \label{sec:previous}}

\subsection{Code-switching, Code-mixing \label{ssec:linguistics}}
Code-switching (CS) can be defined as the alternate use of two languages within the same utterance and can happen either between sentences (inter-sentential CS) or within a sentence (intra-sentential CS or \emph{code-mixing})~\citep{gumperz1982discourse}. While loanwords are often seen as a simple form of CS, their assimilation into a foreign linguistic system sometimes yields a mixed use of languages \emph{within a single word}. For the purpose of this work, we mostly focus on inter-sentential CS and use the terms code-switching and code-mixing interchangeably, even though our approach could in fact apply to longer chunks of texts. From an abstract perspective, the main trait of CS is thus the juxtaposition of  two (or more) languages within a single segments, a view that is also adopted in e.g.\ from \citet{bali-etal-2014-borrowing}. From this perspective, CS ID can be formulated as identifying more than one language ID in a given text segment. We also use the fact that mixing does not take place randomly~\citep{myers1997duelling}, and that one language plays a dominant role and provides the linguistic structure into which inserts from other languages can take place.   

In the next paragraph, we discuss two previous approaches that share this view and which serve as the foundation of \genericname.  For other related works, refer to~\Appref{related}.

\subsection{Detecting CS with Lexical Anchors}
Our work is most closely related to the research of \citet{mendels-etal-2018-collecting}. They propose a method to identify CS data in sentences written in two languages L1 and L2. Their approach first requires a language identifier that is able to label the majority language of a document as language L1, even when the document also contains words that belong to L2. This aligns with our setup, as sentence-level LID models trained on monolingual texts often demonstrate similar performance on CS data, primarily predicting the dominant language L1~\citep{burchell-etal-2024-code}.

\citet{mendels-etal-2018-collecting} also introduce the concept of  \emph{anchors} for each language, defining an anchor as a word belonging to only one language within a language pool \(\mathbb{L}\). The set of anchors in their work is computed based on the analysis of monolingual corpora, and constitutes an external resource to their CS LID system. To relax the definition of anchors, they also introduce the notion of \emph{weak anchor} for a language L2 relative to some other language L1: an anchor is considered a weak anchor' if it is observed in monolingual L2 corpora but not in monolingual L1 corpora.

In their definition of CS  for L1+L2 sentences, a sentence is then considered CS if and only if it is predicted to be in language L1 by the LID model and contains at least one weak anchor from the L2 anchor set (relative to L1). Our method shares similarity with this work in that, for L1+L2 sentences, the initial step consists in the identification of L1. However, while their approach requires the identification of sets of weak anchors for each language pair, we identify the minority language(s) L2 using only features that are internal to the main LID model, dispensing from the need to compile external resources.

\subsection{CS Detection as Set Prediction Problem}

Another work that is closely related to ours is the research conducted by \citet{burchell-etal-2024-code}. They use three different sentence-level LID models for CS LID: 1) OpenLID~\citep{burchell-etal-2023-open}, a high-quality LID model operating at the sentence level; 2) Multi-label OpenLID, which is similar to OpenLID but is trained with a binary cross-entropy loss instead of the conventional cross-entropy, and delivers Yes-No decisions for each possible language;\footnote{See \fasttext documentation: \\\href{https://fasttext.cc/docs/en/supervised-tutorial.html\#multi-label-classification}{\path{fasttext.cc/docs/en/supervised-tutorial.html#multi-label-classification}}.} and 3) Franc~\citep{franc}, which uses trigram distributions in the input text and a language model to compute languages and their scores.

However, the result of these models on CS LID are not very promising especially for the Turkish-English CS dataset (see~\Secref{results}). One reason is that the occurrence of one single English word in a Turkish sentence is tagged in the gold reference as an instance of CS. Yet, one single word may not be enough to yield large logit values for the English label in these difficult predictions. But this is not the only reason these models fail. Scaling the baseline LID to support more languages, which is a strong motivation behind models such as GlotLID~\citep{kargaran-etal-2023-glotlid} and OpenLID, makes CS LID predictions more challenging. For instance, when the model encounters a Turkish-English sentence and predicts Turkish as the top language, the second best prediction may not be English, but a language closest to Turkish instead, such as North Azerbaijani or Turkmen, which have more active ngram features in the CS sentence than English.
Consider, for instance, the example sentence from \citet[Table~9]{burchell-etal-2024-code}:

\begin{verbatim}
    bir kahve dükkanında geçen film 
    tadında güzel bir şarkıya ayrılsın 
    gece falling in love at a coffee shop
\end{verbatim}

OpenLID's top 5 predictions for this sentence are: 1) Turkish, 2) North Azerbaijani, 3) Crimean Tatar, 4) Turkmen, 5) Tosk Albanian, with English predicted as the 15th most likely language. Yet, for a speaker of either Turkish or English, it is obvious that this sentence is a mixture of just these two languages.  To solve this, \genericname suggests to mask the Turkish part of the sentence:

\begin{verbatim}
    <MASK> film <MASK>
    falling in love at a coffee shop.
\end{verbatim}

If we now ask OpenLID to predict this masked sentence (without the token \verb|<MASK>|), the top prediction would be English with 0.9999 confidence. \genericname makes models such as OpenLID much more suitable for this task. Details on how \genericname computes the masked parts are in~\Secref{method}.

\section{\genericname}\seclabel{method}

\subsection{\fasttext-based LIDs}

In this paper, we explore the use of \genericname for LIDs based on the \fasttext~\citep{bojanowski-etal-2017-enriching} architecture. However, it is also possible to apply \genericname to other LIDs, as long as they enable to determine how much each feature (\eg word) contributes to each supported language. \fasttext is one of the most popular LID architectures due to its open-source nature, high performance, ease of use, and efficiency. \fasttext classifier is a multinomial logistic classifier that represents the input sentence as a set of feature embeddings, making it easy to assess each feature's contribution to the final prediction.

Given a sentence $s$, let $f_1, f_2, \ldots, f_T$ represent
the features extracted from $s$. Note that these features are linearly ordered,
i.e., $f_i$ precedes $f_{i+1}$ in $s$. 
\fasttext maps these features onto vectors in $\mathbb{R}^d$ via feature embeddings $\mathbf{x}_1, \mathbf{x}_2, \ldots, \mathbf{x}_T$. The dimensionality of these embeddings, denoted $d$, is a hyperparameter. A base LID using \fasttext{} architecture computes the posterior probability for a language $c \in [1:N]$ by applying the softmax function over logits as:

\begin{equation}\eqlabel{predict}
P(c|s) = \frac{\exp( \mathbf{b}_{c} \cdot \frac{1}{T} \sum_{t=1}^{T} \mathbf{x}_t)}{\sum_{c'=1}^{N} \exp(\mathbf{b}_{c'} \cdot \frac{1}{T} \sum_{t=1}^{T} \mathbf{x}_t)}.
\end{equation}

\(P(c|s)\) is the base LID probability of the input text $s$ belonging to language \(c\), 
\(\mathbf{b}_c\) is the weight vector for language \(c\), and \(N\) is the total number of classes supported by the base LID.

To evaluate how much each feature contributes to each supported language, we need to compute logits separately for each feature. For simplicity and alignment with the \fasttext tokenizer (which considers white-spaces as token boundaries), we set the level of granularity of features to be the word level. The word-level feature embedding is obtained as the summation of all feature embeddings that build each word. Noting $W$ the number of words in a sentence $s$, we define the $N\times W$ matrix \(\mathbf{V}(s)\), where each element \(\mathbf{V}_{c,t}(s)\) represents the logits for language \(c\) and word-level feature \(\mathbf{x}_t\):

\begin{equation}\eqlabel{matrix_v}
\mathbf{V}_{c,t}(s) = \mathbf{b}_{c} \cdot \mathbf{x}_t.
\end{equation}

\subsection{The \genericname Method \seclabel{model}}

We define the \genericname algorithm in alignment with \citet{burchell-etal-2024-code}: given an input text, the objective is to return a set of codes corresponding to the language(s) it contains. However, \genericname is more explainable and provides insights into which parts of the sentence contributed to its decision. The \genericname algorithm works as follows:

\textbf{Input:}
\begin{itemize}[topsep=0pt,partopsep=0pt,parsep=0pt,itemsep=0pt]
\item[1)] sentence $s$.  
\item[2)] $\alpha$, an integer parameter used to define \emph{strong associations} between words and languages: having a language appear in the top-$\alpha$ logit values for a word is a strong cue that this word belongs to that language. 
\item[3)] $\beta$, an integer parameter used to define \emph{weak associations} between words and languages: languages appearing in the top-$\beta$ logit values for a word are weakly associated with that word.  $\beta$ is always greater than $\alpha$.
\item[4)] $\tau$, a threshold representing the minimum size of a sentence (in bytes) for which the LID makes reliable decisions.  
\item[5)] $\lambda$, a parameter defining the number of times the algorithm should be repeated.
\end{itemize}
\newpage
\textbf{Output:}
\begin{itemize}[topsep=0pt,partopsep=0pt,parsep=0pt,itemsep=0pt]
\item[1)] List of predicted languages, along with their associated word-level features.
\end{itemize}
\textbf{Procedure:}
\begin{itemize}[topsep=0pt,partopsep=0pt,parsep=0pt,itemsep=0pt]
\item[0)] Take sentence $s$ and compute $\mathbf{V}(s)$ using \eqref{matrix_v}. Assign $s$ to variable $u$.
\item[1)] Compute the posterior probability for each possible language using~\eqref{predict}. Find the most likely class (\(L1 = \arg\max_c P(c|u)\)) along with its corresponding probability \(P(L1|u)\). Assign L1 to variable $\text{L}_u$.
\item[2)] Process column \(\mathbf{V}_{:,t}(s)\) for each unmasked word \(t\) in $u$. If the value of \(\mathbf{V}_{\text{L}_u,t}(s)\) is in the top-\(\beta\) values for that column, then assign word \(t\) to language $\text{L}_u$.  If the value of \(\mathbf{V}_{\text{L}_u,t}\) is among the top-\(\alpha\) values for that column, mask word $t$ from sentence $u$.

Masked words play here a role similar to the anchors used in \citep{mendels-etal-2018-collecting}: recall that for these authors, anchor words are selected to uniquely identify one language -- there removal is likely to decrease the recognition of L1, without impacting the ability to recognize L2. In our approach, we identify these \emph{pseudo-anchors} on the spot, relying on the LID internal scoring procedure.

\item[3)] check if length of $u$ (in bytes, ignoring masked words) is greater than $\tau$. If not, then terminate. This is one termination condition (for additional considerations, refer to~\Appref{consideration}). Setting $\tau =0$ will just check that the masked sentence is not empty, but it is better to use a non-zero threshold, as most sentence-level LIDs do not reliably predict short sentences \cite{jauhiainen2019automatic}. 
\item[4)] if the number of iterations is lower than $\lambda$ then go to back to step~1, else stop.
\end{itemize}

The complexity of this greedy procedure is $O(\lambda\times{} T\times{} N\log\beta)$.

\section{Experiments and Results \seclabel{results}}

Here, we provide an overview of our baselines and test data. We assess the performance of the baselines by testing them both with and without \genericname. Our setting of hyperparameters is explained in~\Appref{hyperparameters}.

\subsection{Baselines}
Our baseline LID models are OpenLID\footnote{\url{https://huggingface.co/laurievb/openlid}} (supporting $\approx$200 languages) and GlotLID v3.0\footnote{\url{https://huggingface.co/cis-lmu/glotlid}} (supporting $\approx$2100 languages), two LIDs based on the \fasttext architecture. For a fair comparison between these models, we limit the languages that GlotLID supports to the same set as OpenLID (see details in \Appref{limitlid}). Two exceptions are romanized Nepali (nep\_Latn) and Hindi (hin\_Latn), which are not supported by OpenLID, but for which we also have test data that is also used to evaluate \genericname with GlotLID.

\subsection{Test Data}

We choose Turkish-English~\citep{yirmibesoglu-eryigit-2018-detecting}, Hindi-English~\citep{aguilar-etal-2020-lince}, Nepali-English~\citep{aguilar-etal-2020-lince} and Basque-Spanish~\citep{aguirre-etal-2022-basco}, as our test datasets. We have data for four CS labels and six single labels (see~\tabref{results}). Details regarding these test sets, preprocessing, their descriptions, and information on access are in \Appref{data-selection}.

\subsection{Metrics}

We use the number of exact (\#EM) and partial matches (\#PM), along with the count of false positives (\#FP) as the main metrics in our evaluation. To ensure clarity and prevent misinterpretation of the results, we report the absolute number of instances rather than percentages.
\begin{itemize}[topsep=0pt,partopsep=0pt,parsep=0pt,itemsep=0pt]
\item[1)] \#EM: This metric counts a prediction as a match when it exactly matches the true 
\item[2)] \#PM: This metric counts a prediction as a match when only part of the information is correct: for a single label, if it is part of the prediction; for a CS label, if part of the label exactly matches the prediction. 
\item[3)] \#FP: If any label other than X is misclassified as X, it counts as an FP for X. We do not consider the \#FP for single labels, as partial matches of CS are counted as FP for single labels. Therefore, we only report the FP for CS sentences.
\end{itemize}

\begin{table*}[h]
    \centering
    \small
    \resizebox{1.0\linewidth}{!}{
    \begin{tabular}{lllllllllll}
    & &  \multicolumn{4}{c}{Baseline + \genericname} &  
    \multicolumn{4}{c}{Baseline} & \\
    \cmidrule(lr){3-6} \cmidrule(lr){7-10}
    && 
    \multicolumn{2}{c}{\#EM/\#PM $\uparrow$}
    & \multicolumn{2}{c}{\#FP $\downarrow$}
    & \multicolumn{2}{c}{\#EM/\#PM $\uparrow$} 
    & \multicolumn{2}{c}{\# FP $\downarrow$}
    \\
     & \#S & GlotLID & OpenLID & GlotLID & OpenLID & GlotLID & OpenLID & GlotLID & OpenLID \\
    \cmidrule(lr){3-4} \cmidrule(lr){5-6} \cmidrule(lr){7-8} \cmidrule(lr){9-10}
    CS Turkish--English & 333 & \textbf{91}/328 & \underline{68}/327 & 0 & 0 & 4/327 & 4/326 & 0 & 0 \\ 
    CS Basque--Spanish & 440 & \underline{43}/430 & \textbf{47}/426 & 0 & 0 & 9/426 & 9/424 & 0 & 3 (from Spanish) \\
    CS Hindi--English & 253 & \textbf{29}/219 & - & 0 & - & \underline{5}/211 & - & 0 & - \\
    CS Nepali--English & 712 & \textbf{22}/444 & - & 0 & - & \underline{0}/420 & - & 0 & - \\
    Single Basque & 357 & 354/354 & 355/355 & - & - & 353/353 & 355/355 & - & - \\
    Single Spanish & 347 & 335/337 & 297/300 & - & - & 337/340 & 287/311 & -& - \\ 
    Single Turkish & 340 & 333/337 & 329/334 & - & - & 335/337 & 329/335 & - & - \\
    Single Hindi & 29 & 18/19 & - & - & - & 17/18 & - & - & - \\
    Single Nepali & 197 & 63/75 & - & - & - &  68/72 & - & - & - \\
    Single English & 508 & 459/490 & 428/469 & - & - & 486/490 & 455/462 & - & - \\
    \end{tabular}
    }
    \caption{Number of exact (\#EM) and partial matches (\#PM) and count of false positives (\#FP) calculated over CS and single label test instances. The best exact match for CS instances is in bold, and the second is underlined. \#S reports the number of sentences for each test set.}
    \tablabel{results}
\end{table*}

\subsection{Results}

\tabref{results} presents the results on the test data for two baseline LIDs and two settings, with and without \genericname. The best exact match (\#EM) for CS labels is in boldface in the table, demonstrating that the baseline with \genericname{} achieves better performance compared to the baseline without it. Partial matches (\#PM) in both settings (with and without \genericname) are quite similar.

For CS Turkish-English, \genericname detects 91 CS at best, compared to 4 without it. For Basque-Spanish, \genericname detects 47 CS, versus 9 without it. For Hindi-English, \genericname detects 29 CS, compared to 5 without it. For Nepali-English, \genericname detects 22 CS, while none are detected without it.

In all single-language test instances, GlotLID outperforms OpenLID. This is also the case for CS language instances, except for Basque-Spanish. Considering the relatively poorer performance of OpenLID in both single Basque and single Spanish, overall, GlotLID proves to be the better model for these tasks.

\textbf{Additional Considerations.}
For {CS instances}:
1) The difference between \#PM and \#EM corresponds to the number of times only one of two mixed languages in a CS instance is predicted.
2) The difference between number of sentences (\#S) and \#PM corresponds to the number of times none of the languages in the CS instance is predicted.
In all CS setups, the \#EM and \#PM value in the baseline with \genericname{} are always greater than without. Additionally, the difference between \#PM and \#EM is also smaller, which indicates a higher precision in CS LID.

For {single language instances}:
1) The difference between \#PM and \#EM corresponds to the number of times the single label instance is classified as part of a multi-label instance.
2) The difference between \#S and \#PM corresponds to the number of times a single label is never predicted, even as part of a multi-label instance.
For all single language instances, the results are quite similar except for single English, where the number of incorrect CS in baseline with MaskLID (\#PM - \#EM) is greater than with baseline alone. To address this, using a larger minimum length $\tau$ helps decrease the number of CS false positives. For single English, in GlotLID with \genericnamenospace{} setting, increasing $\tau$ from $20$ to $25$ raises the \#EM from 459 to 473; however, it reduces the \#EM in GlotLID with \genericnamenospace{} setting for CS Turkish-English from 91 to 67, CS Hindi-English from 29 to 26, and CS Nepali-English from 22 to 18. Examples of successes and failures of \genericname{} are provided in \Appref{examples}.

\section{Conclusion}

We present \genericname, a simple, yet effective, method for scalable code-switching (CS) language identification (LID). \genericname is designed to complement existing high-performance sentence-level LID models and does not require any training. In our experiments, \genericname increases CS LID by a factor of 22 in Turkish-English, by 22 in Nepali-English, by 6 in Hindi-English and by 5 in Basque–Spanish.

In future work, we aim to explore the use of subword-level, instead of word-level features, extending the applicability of the method to languages that do not use spaces for word separation. Additionally, we plan to generalize this method to other LID models using techniques like LIME~\citep{ribeiro2016should} to map features to languages. Last, we intend to apply \genericname on the web data, in the hope that it will help build larger high-quality web corpora for CS.

\section*{Limitations}

The CS testsets we use in this study only represent a small subset of potential uses of CS languages. Creating additional CS datasets for more languages would definitely be an extension of this work. \genericname uses hyperparameters, and changing the model and the set of languages it supports may require adjustments to these parameters. Although \genericname detects more CS than the standalone baseline LID models, it still has a long way to go to predict the majority of them. One important source of remaining errors is loan words, where the L2 insert is just one word long: these cannot be detected with out current hyperparameter settings. The performance of \genericname is also bound by the LID it uses; it might not have good performance for some languages, resulting e.g.\ in a large number of false positives.

\section*{Ethics Statement}

\genericname uses openly available open-source LID models and does not require any additional resources except for hyperparameters. Concerning the evaluation data, these datasets have undergone anonymization to safeguard the privacy of all parties involved. We provide links to the data and do not host it ourselves. We provide detailed descriptions of our method and evaluation process. Additionally, we make our code openly available to foster collaboration and reproducibility.

\section*{Acknowledgements}
The authors thank the anonymous reviewers and editors for their comments of the previous version of this work. This research was supported by DFG (grant SCHU 2246/14-1).

% Bibliography entries for the entire Anthology, followed by custom entries
%\bibliography{anthology,custom}
% Custom bibliography entries only
\bibliography{main}

\appendix

\section{Related Work}\seclabel{related}
LID has been a longstanding and active research area in NLP~\citep{jauhiainen2019automatic}. Past research in LID can be classified into two primary subcategories: 1) monolingual LID; 2) CS LID.

The first category is designed under the assumption that the text is entirely monolingual, or the text contains discrete monolingual chunks (\eg sentences) in different languages. The aim of these works is to identify the language of the whole text or each chunk. The majority of research on this topic has been focused on covering more languages, with recent work claiming to cover over a thousand~\citep{kargaran-etal-2023-glotlid,adebara-etal-2022-afrolid,nllbteam2022language,burchell-etal-2023-open,dunn2020mapping, dunn-edwards-brown-2024-geographically-informed, jauhiainen-etal-2022-heli, brown2012finding}.

The second category has received less attention than the first category. LID at either the document or sentence level is not effective in accurately identifying CS, which may occur within a sentence. LIDs that identify languages at the word level are proposed to address this issue. The majority of studies have focused on scenarios where two predefined languages are looked for in the input, specifically concentrating on binary language detection at the word level~\citep{nguyen-dogruoz-2013-word,das-gamback-2014-identifying,elfardy2013code, king-abney-2013-labeling, al-badrashiny-diab-2016-lili}. While some attempts choose sentence-level granularity \citep{stensby2010language, lavergne-etal-2014-automatic}, most CS LIDs prefer operating at the word or token level. Nevertheless, certain approaches broaden the analysis to the character level~\citep{kocmi-bojar-2017-lanidenn}. Among the most recent works on CS LID, \citet{kevers-2022-coswid} propose a method to locate CS, primarily in multilingual documents when language diversity is unstructured. It uses a sliding window and determines the local language of each token. This method requires linguistic resources such as word lists and monolingual corpora. \citet{rijhwani-etal-2017-estimating} acknowledge the challenges in building word-level LID for CS LID. They propose an unsupervised word-level LID approach and apply it to estimate language pairs code-switched on Twitter. Their findings indicate that approximately 3.5\% of tweets were code-switched.
\citet{mager-etal-2019-subword} extend the LID task from the word level to the subword level, involving the splitting of mixed words and tagging each part with an LID. However, training such LID models at the subword level requires CS training data, which is not practical on a larger scale.

\section{Confidence in \genericname}\seclabel{consideration}

We discuss here additional considerations regarding the design \genericname, notably aimed the keeping a good balance between over and under detection of labels, which is a key aspect to reliably detect instances of CS.

A first comment is that in our approach, the value of parameter \(\alpha\) is kept constant. An extension would vary this value during iterations, depending on the desired level of CS-sensitive results. However, selecting a smaller \(\alpha\) increases the likelihood of a language being chosen again in the next round(s). In such cases, the \(\alpha\) value for the next round should be increased so that more words belonging to L1 are masked.

To ensure that \genericname yields a low false positive rate (FPR), the feature set assigned to language $\text{L}_u$ in step~2 should have a minimum length (in byte) $\tau$. If not, we should increase the \(\beta\) value and repeat the process again to obtain a larger feature set, and evaluate whether the confidence probability prediction for this set is high. If not, terminate the procedure. It is important to note that \(\beta\) does not play a role in masking, as only \(\alpha\) affects this process. The reason for defining both \(\alpha\) and \(\beta\) instead of relying solely on \(\alpha\) is to ensure a minimum byte size so that the probability prediction for this feature set can be trusted and to guarantee its high confidence. Typical \(\alpha\) values should thus be lower than \(\beta\) and only target the features that strongly cue language and should accordingly be masked.

Maintaining high confidence in steps 1 and 4 is more tricky; the reason for the low confidence probability in these steps could be the presence of another language. However, it could also be because the text is not among the languages supported by the LID~\citep{kargaran-etal-2023-glotlid}. We suggest using a low confidence threshold for these steps or not using one at all.

Finally, our algorithm uses two termination conditions, one based on the minimum sentence length ($\tau$) , one based on the maximum number of languages in a given sentence ($\lambda$): 2 or 3 is recommended. In our test dataset, we know in advance that the number of languages is at most $2$.

\section{Experimental Settings}\seclabel{experiment-details}

\subsection{The Label Sets of LIDs}\seclabel{limitlid}

Following the labeling proposed by \citet{nllbteam2022language}, our two baseline LIDs use language-scripts as labels. They define a language-script as a combination of a ISO 639-3 language code and a ISO 15924 script code.

We constrain GlotLID to the set of languages supported by OpenLID. Most of the labels supported by OpenLID are supported by GlotLID. The total number of labels is 201 for OpenLID, and we select 200 labels for the constrained version of GlotLID. The only difference is due to the fact that OpenLID uses two labels for the Chinese language (zho), written in Hans and Hant scripts, whereas GlotLID combines both under the label Hani. Also, GlotLID does not support acq\_Arab, nor does it not support labels pes\_Arab and prs\_Arab individually (as OpenLID does) but as the merged macrolanguage fas\_Arab. To compensate for the lack of these two labels and to also perform experiments for Hindi and Nepali in romanized script, we add hin\_Latn and npi\_Latn to the set of labels for constrained GlotLID.

To restrict a \fasttext-based LID model to a specific subset of languages, as indicated by \eqref{predict}, we only need to consider the \(\mathbf{b}_c\) values for languages \(c\) that are members of the chosen set of languages. This implies that languages not included in this set will be excluded from the softmax computation. Additionally, the rows belonging to these languages are also deleted from the matrix \(\mathbf{V}(s)\) (\eqref{matrix_v}).

\subsection{Hyperparameters}\seclabel{hyperparameters}

We here explain the hyperparameters specific to each method.

\textbf{\genericname.} We generated 12 small synthetic code-switch corpora by combining sentence parts from French, English, Arabic, and Persian languages, ensuring a presence of at least 30\% from each of the two languages participating in the final sentence. Subsequently, we applied \genericname with different hyperparameters to achieve the best results. The hyperparameters derived from this method, which we used for the experiments in this paper, are as follows: $\alpha = 3$, $\beta = 15$, $\lambda = 2$, and $\tau = 20$. Additionally, we employed a high-confidence threshold of $0.9$ for OpenLID and GlotLID to evaluate the probability predictions for the feature set in step 2 of the algorithm, as further detailed in \Secref{consideration}.

\textbf{Baseline.}
Following \citet{burchell-etal-2024-code}, we use a threshold of 0.3 to select languages (\ie among all languages supported by the model, the languages with confidence probability greater than 0.3 are selected). However, for a fairer comparison (since $\lambda = 2$), we only consider the top two that pass this threshold.

\section{Data Selection}\seclabel{data-selection}

The CS test sets available for consideration cover a small potential language set \citep{jose2020survey, aguilar-etal-2020-lince}. 
Accessing suitable CS test sets for evaluating our method poses several challenges:

1) Arabic dialects, such as Standard Arabic-Egyptian Arabic, are represented in some CS datasets~\citep{elfardy2013code, aguilar-etal-2020-lince}. However, none of the baseline LID models yield impressive results for Arabic dialects. For instance, according to \citet[Table~3]{burchell-etal-2024-code}, OpenLID exhibits the worst FPR for Standard Arabic and Egyptian Arabic among all the languages it supports.

2) Certain datasets present unrealistic scenarios for testing our method. For example, Mandarin-English datasets with Mandarin written in Hani script and English in Latin script~\citep{lovenia-etal-2022-ascend}. Methods employing script detection can separate perfectly Hani from Latin, and perform two separate LID predictions.\footnote{For example, GlotScript~\citep{kargaran-etal-2024-glotscript-resource} provides a \texttt{separate\_script} function that divides text based on different scripts: \href{https://github.com/cisnlp/GlotScript}{\path{github.com/cisnlp/GlotScript}}.} This does not showcase the advantages of \genericname and the performance only is dependent to the LID performance.

3) Many accessible datasets involve CS between one language and English. 

Given these challenges, we decided to use datasets involving English in three sets (Turkish-English, Hindi-English, Nepali-English) and another set with CS between languages without English (Basque-Spanish). The Turkish-English and Basque-Spanish datasets are also used by \citet{burchell-etal-2024-code}. We use the code provided by these authors to label them into sentence-level tags. 

\textbf{Turkish-English.}
\citet{yirmibesoglu-eryigit-2018-detecting} developed a Turkish–English dataset for CS as part of their work on CS LID for this langauge pair. The dataset is sourced from Twitter and the Ekşi Sözlük online forum. Labels in this dataset are assigned at the token level, indicating whether each token is Turkish or English. The dataset comprises 376 lines of data, and 372 of these sentences are labeled as CS. However, for our purposes, we also require monolingual datasets in these languages, not just CS data. To address this, we created a monolingual version of the CS data for the Turkish language by removing tokens labeled as English. A similar approach cannot be applied to create an English monolingual dataset, as the English parts of the data are short sentences and would adversely impact the quality of the English monolingual data. The original dataset can be found here: \href{https://github.com/zeynepyirmibes/code-switching-tr-en}{\path{github.com/zeynepyirmibes/code-switching-tr-en}}.

\textbf{Basque-Spanish.}
The Basque–Spanish corpus~\citep{aguirre-etal-2022-basco} comprises Spanish and Basque sentences sourced from a collection of text samples used in training bilingual chatbots. Volunteers were presented with these sentences and tasked with providing a realistic alternative text with the same meaning in Basque–Spanish CS. The dataset consists of 2304 lines of data, with 1377 sentences labeled as CS, 449 as Basque, and 478 as Spanish. The original dataset is available at: \href{https://github.com/Vicomtech/BaSCo-Corpus}{\path{github.com/Vicomtech/BaSCo-Corpus}}.

\textbf{Hindi-English \& Nepali-English.} \citet{aguilar-etal-2020-lince}
provide a benchmark for linguistic CS evaluation,
used in previous shared tasks on CS LID  \citep{solorio-etal-2014-overview, molina-etal-2016-overview}. We test on two
of its language pairs, Hindi–
English and Nepali-English, using the validation sets since the test sets are private. These datasets are both sourced from Twitter and are annotated at the word level. The Hindi-English dataset has 739 lines: 322 CS, 31 Hindi, and 386 English sentences. The Nepali-English dataset has 1332 lines: 943 CS, 217 Nepali, and 172 English sentences. We consider both CS and monolingual data for experiments.

\textbf{Preprocessing} Sentence-level LIDs may not perform well on very short sentences. In the corpus creation pipelines using these LIDs, shorter sentences are typically discarded. Therefore, we filter sentences with a length of 20 byte or fewer for monolingual sentences and sentences with a length of 40 byte or fewer for CS sentences. The remaining number of sentences (\#S) for each portion of the data is detailed in \tabref{results}. In addition, we clean user tags and emojis from the datasets before applying LIDs.

\section{Examples}\seclabel{examples}

We showcase below some failed and successful examples of \genericname.

\textbf{Failed Example.} In this example, the only English word is ``status''.
\begin{Verbatim}[commandchars=\\\{\}]
    yarın bir \textcolor{red}{status} yapıp
    işlerin üstünden geçelim
\end{Verbatim}

As we define the minimum length for each selected language to be at least $\tau = 20$ byte, this sentence gets classified as Turkish, which is acceptable. If, otherwise, ``status'' would be evaluated alone, OpenLID would predict ``Norwegian Nynorsk'' language, and GlotLID ``Kinyarwanda''. This is the reason why $\tau$ is important to be set because otherwise the result of LID cannot be trusted. The average length of the English part of sentences in the CS Turkish-English getting classified solely as Turkish by GlotLID + MaskLID is 17.858 bytes and by OpenLID + MaskLID is 19.877 bytes. So the main reason for failing these models here is the English part of this sentences is short and often does not pass the minimum length condition.

\textbf{Successful Example.} In this example, ``deadline crash walking I heard it at study'' are the English words inserted in the Turkish sentence. These words are not next one to the other, so methods that only consider sliding windows might fail. \genericname does not depend on the position of words in a sentence and correctly classify this example as Turkish-English CS. 

\begin{Verbatim}[commandchars=\\\{\}]
    ya \textcolor{red}{deadline} gelmişti çok büyük
    bir \textcolor{red}{crash} olmuş arkadaşlarla
    \textcolor{red}{walking} yaparken \textcolor{red}{I heard it at}
    boğaziçi sesli \textcolor{red}{study}
\end{Verbatim}

However, predicting it using solely based on OpenLID results in the top 3 labels being ``Turkish'', ``Turkmen'', and ``North Azerbaijani''. The average length of the English part of sentences from CS Turkish-English getting classified correctly as CS Turkish-English by GlotLID + MaskLID is 42.121 bytes and by OpenLID + MaskLID is 45.294 bytes.

\end{document}